\documentclass[runningheads]{llncs}
\usepackage[T1]{fontenc}
\usepackage{graphicx}
\begin{document}
\title{TalentCLEF at CLEF2026: Skill and Job Title Intelligence for Human Capital Management}
\titlerunning{TalentCLEF at CLEF2026}
\author{Luis Gasco\inst{1}\orcidID{0000-0002-4976-9879} \and
Hermenegildo Fabregat\inst{1} \and
Laura García-Sardiña\inst{1} \and
Paula Estrella\inst{1} \and
Casimiro Pio Carrino\inst{1} \and
Daniel Deniz\inst{1} \and
Alvaro Rodrigo\inst{2} \and
Rabih Zbib\inst{1}
}
\authorrunning{L. Gasco et al.}
%
\institute{Avature Machine Learning, Spain \\
\email{machinelearning@avature.net} \and
NLP \& IR Group at UNED, Madrid, Spain
}
\maketitle              
\begin{abstract}
This paper presents the second edition of the TalentCLEF Challenge, which will run as an evaluation lab part of CLEF2026. The aim of TalentCLEF is to promote the development of systems and methods that use Natural Language Processing (NLP) in the field of Human Capital Management (HCM), fostering approaches that ensure fairness in results, operate across multiple languages, and adapt to diverse industries. To this end, TalentCLEF establishes public benchmarks where research teams can compare methods and share findings, moving the field toward more practical and impactful NLP solutions that effectively address the real needs of workforce management.

This year's lab will feature two tasks designed to foster the development and evaluation of systems that support key HCM activities such as talent matching, upskilling, reskilling, and skill gap detection: (i) Task A – Contextualized Job-Person Matching, focused on retrieving and ranking suitable candidates for specific job positions using context-rich and privacy-preserving data, and (ii) Task B – Job-Skill Matching with Skill Type Classification, centered on identifying relevant skills for a given job title and classifying them by their type within the job profile.

TalentCLEF website: \url{https://talentclef.github.io/talentclef/}

\keywords{Natural Language Processing  \and 
Human Capital Management  \and 
Human Resources  \and 
Multilinguality  \and 
Cross-linguality  \and 
Skill Predictions  \and 
Job Title Ranking}
\end{abstract}
\section{Introduction}

Over the past decade, technology has significantly transformed the labor market. Digitalization, the expansion of remote work, and the growing internationalization of organizations have shaped an increasingly global, multilingual, and competitive environment, both for companies seeking to attract talent and for workers accessing new professional opportunities. In this context, the emergence of Artificial Intelligence (AI) is introducing an even more significant transformation, affecting both the nature of jobs and the competencies required to perform them. According to recent projections, more than 1.1 billion jobs could be transformed by the end of the decade~\cite{wef2023reskilling}, with approximately 70\% of the skills currently used in most occupations expected to change as a result of technological evolution~\cite{linkedin2025workchange}. This trend is already evident: In 2024, 74\% of employers reported difficulties in finding candidates with suitable skills for their job openings~\cite{manpower2024talentshortage}.

This transformation process has profound implications for Human Capital Management (HCM), particularly in key areas such as recruitment, learning, and career management. Identifying and selecting suitable candidates has become increasingly complex in an environment where professional requirements evolve very fast. Both organizations and employees require effective tools to keep developing new professional capabilities, as well as mechanisms that support flexible and adaptable career paths within a constantly changing labor market. Companies must attract and manage the right talent, and workers need to update and show their capabilities effectively in a global and highly competitive context.

In recent years, the application of Natural Language Processing (NLP) technologies to HCM has advanced substantially. Various initiatives~\cite{bogers2025fifth,hruschka2024proceedings,gasco2025talentclef} have contributed to the consolidation of a research community focused on the use of language models and machine learning to represent and analyze information related to skills, occupations, and career paths. These developments have enabled the creation of systems that can extract and normalize relevant information from semi-structured records like résumés and job descriptions~\cite{senger2024deep,retyk2023r,zhang2022skill,zhang2022skillspan,garcia2023normalisation,retyk2024melo}, as well as measuring semantic similarity between profiles and job postings~\cite{otani2024natural,zbib2022learning,decorte2021jobbert,deniz2024combined}. At the same time, recent research has emphasized the importance of addressing aspects such as detection of algorithmic bias and fairness evaluation in automated decision-making processes within the labor context~\cite{albaroudi2024comprehensive,garcia2025measuring}.

Despite these advances, the development of AI systems applied to talent management continues to face structural limitations. Most studies rely on private data, which restricts reproducibility and hinders the comparison of different approaches. The few public resources available often lack consistent annotation criteria or sufficient information regarding their quality. Standardized evaluation frameworks for assessing system performance and fairness are still missing. As a result, comparing models and methods remains challenging, and research in this area progresses in a fragmented manner.

TalentCLEF\footnote{https://talentclef.github.io/talentclef/} was conceived precisely to address this gap~\cite{gasco2025overview,gasco2025talentclef}. The strong participation in the first TalentCLEF edition, which attracted 76 teams and generated 280 system submissions, confirmed the growing interest of the community in this domain. Building on this success, a second edition has been launched under the umbrella of CLEF 2026 to consolidate and expand the initiative. Its goal is to continue providing an open, reproducible and multilingual evaluation environment that promotes the development and comparison of AI technologies applied to talent management.

\section{TalentCLEF 2026 Evaluation Lab}
The second edition of TalentCLEF builds on the foundations established in 2025~\cite{gasco2025brief}. The new lab consists of two complementary tasks that operate on realistic multilingual synthetic data and focus on two key entity types in the employment domain: job titles and skills. Unlike last year’s edition, which centered on shorter and non-contextual inputs, this year’s edition extends that foundation by incorporating richer contextual information from synthetic job descriptions and résumés. Specifically, Task~A addresses multilingual job–person matching, while Task~B continues with the same focus, predicting skills relevant to specific job positions, two challenges that are central not only to recruiting, but also to learning and career development.

\begin{figure}[!ht]
\centering
\includegraphics[width=1\textwidth]{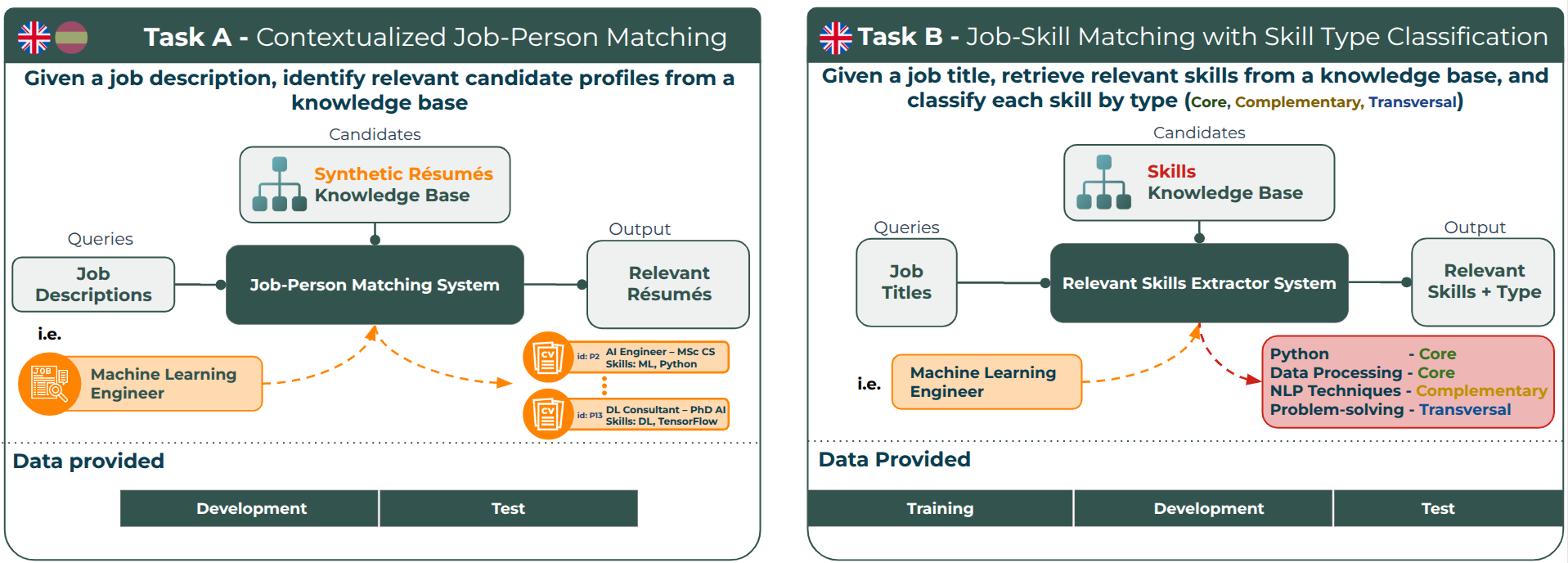}
\caption{Diagrams of TalentCLEF Task A and B} \label{fig2}
\end{figure}

\subsection{Task A – Context-Aware Job-Person Matching}
Matching candidates to job offers is a central problem in recruitment systems. Traditionally, job-candidate matching has relied on the comparison of isolated entities, such as skills or job titles, both extracted from résumés and job descriptions. However, such approaches often overlook the broader semantic and contextual relationships between these elements, largely due to technical and scalability constraints~\cite{zhao2021embedding}. The evolution of Large Language Models (LLMs) is opening new opportunities for context-aware candidate–job matching. These models enable systems to move beyond the similarities of isolated entities towards a deeper, holistic understanding of both candidates and job postings within their full semantic context~\cite{lo2025ai,gan2024application}.

This year, Task A specifically challenges participants to develop systems capable of identifying the most appropriate candidate profiles for a given job offer. In contrast to the previous edition, which focused on measuring the similarity between pairs of job titles to build job recommendation systems, this version shifts the focus toward matching job offers to suitable candidate résumés. These documents are synthetically generated from structured data originally derived from real job descriptions and résumés, ensuring that they are realistic while preserving privacy. As such, there are no privacy risks associated with the data. 

For the task, participants will be provided with a development set and a test set, composed of a total of 100 job descriptions and 300 synthetic résumés. The development set can be used to design, train, and validate systems that take a job description as input and output a list of candidate profiles ranked according to their relevance. Participants are encouraged to explore techniques such as data augmentation, fine-tuning, prompt engineering, information extraction, or representation learning to improve their models’ performance. On the other hand, the test set will be used in the official evaluation phase, which is designed to ensure a fair and consistent comparison across all submitted runs.

\subsection{Task B – Job-Skill Matching with Skill Type Classification}

Employers have shifted the focus of their talent management strategies from role-based to skill-based. Understanding which skills are associated with a job role is essential not only for improving recruitment systems but also for identifying skill gaps within organizations and designing targeted upskilling and reskilling programs, allowing companies to adapt more rapidly to changes in the labor market.

With this in mind, Task B challenges participants to develop systems capable of automatically identifying the most relevant skills for a given job title. This year’s task extends Task B from the previous edition of TalentCLEF, improving the quality of the dataset and introducing an additional layer of information: Skill type classification. Participants must retrieve the skills that best match each job title and determine whether each skill is specific or transversal to that job title.

As in the previous year, the dataset supporting Task B is divided into three subsets designed to train, validate, and benchmark the proposed systems. The training set contains 5,000 job titles linked to their corresponding relevant skills, providing a solid foundation for model development. The development set includes 200 job titles normalized to ESCO terminology and enriched with the new skill type annotations, enabling participants to fine-tune and validate their systems using richer semantic information. Finally, the test set, comprising 500 job titles, will be used in the official evaluation phase, where participants must generate predictions and submit their ranked outputs and skill types for benchmarking.

\subsection{Evaluation}
Tasks A and B will be evaluated on the Codabench platform, using standard information retrieval metrics. Mean Average Precision (MAP) will serve as the official metric for both tasks, and complementary metrics such as Mean Reciprocal Rank (MRR) and Precision@K will be reported.

For Task A, the evaluations will be conducted in two monolingual settings (English and Spanish), as well as in a cross-lingual setting (en–es). In addition, the evaluation will incorporate a fairness assessment to measure model performance with respect to gender bias, using the Rank-Biased Overlap metric (RBO)~\cite{10.1145/1852102.1852106}. This ensures that the analysis not only captures performance, but also addresses fairness and ethical robustness, which are especially relevant in sensitive areas of application like Human Capital Management.

The top two teams in each monolingual evaluation, the best cross-lingual team, and the model with the best bias control will be highlighted during the workshop. For Task B, the two best-performing teams will be recognized. All of these teams will receive a certificate of achievement acknowledging their results.

\section*{Invitation to Participate}
Anyone interested in joining the TalentCLEF challenge can register on the CLEF conference website. After registering, participants will be able to upload their prediction output via the CodaLab competition system.

%
%
%
%
\bibliographystyle{splncs04}  
\bibliography{bibliography}

@misc{wef2023reskilling,
  author       = {{World Economic Forum}},
  title        = {The Reskilling Revolution: 350 Million People Reached with Future-Ready Skills, Education and Jobs},
  howpublished = {\url{https://www.weforum.org/press/2023/01/the-reskilling-revolution-350-million-people-reached-with-future-ready-skills-education-and-jobs/}},
  year         = {2023},
  month        = {January},
  note         = {Press release},
  organization = {World Economic Forum}
}

@misc{linkedin2025workchange,
  title        = {Work Change Report: AI Is Coming to Work},
  author       = {{LinkedIn}},
  year         = {2025},
  month        = {January},
  institution  = {LinkedIn Corporation},
  url          = {https://economicgraph.linkedin.com/content/dam/me/economicgraph/en-us/PDF/Work-Change-Report.pdf}, 
  note         = {Accessed: 2025-05-18}
}

@misc{manpower2024talentshortage,
  title        = {2024 Global Talent Shortage},
  author       = {{ManpowerGroup}},
  year         = {2024},
  institution  = {ManpowerGroup},
  url          = {https://go.manpowergroup.com},
  note         = {Accessed: 2025-05-19}
}

@inproceedings{gasco2025overview,
  title={{Overview of the TalentCLEF 2025: Skill and Job Title Intelligence for Human Capital Management}},
  author={Gasco, Luis and Fabregat, Hermenegildo and Garc{\'\i}a-Sardi{\~n}a, Laura and Estrella, Paula and Deniz, Daniel and Rodrigo, Alvaro and Zbib, Rabih},
  booktitle={International Conference of the Cross-Language Evaluation Forum for European Languages},
  pages={464--485},
  year={2025},
  organization={Springer}
}

@article{gasco2025brief,
  title={Brief Overview of TalentCLEF 2025},
  author={Gasco, Luis and Fabregat, Hermenegildo and Garc{\'\i}a-Sardi{\~n}a, Laura and Estrella, Paula and Deniz, Daniel and Rodrigo, Alvaro and Zbib, Rabih},
  journal={CLEF (Working Notes)},
  volume={4038},
  pages={4388--4391},
  year={2022}
}

@inproceedings{gasco2025talentclef,
  title={{TalentCLEF at CLEF2025: Skill and Job Title Intelligence for Human Capital Management}},
  author={Gasco, Luis and Fabregat, Hermenegildo and Garc{\'\i}a-Sardi{\~n}a, Laura and Deniz, Daniel and Rodrigo, Alvaro and Estrella, Paula and Zbib, Rabih},
  booktitle={European Conference on Information Retrieval},
  pages={479--486},
  year={2025},
  organization={Springer}
}

@article{zbib2022learning,
  author       = {Rabih Zbib and
                  Lucas Lacasa Alvarez and
                  Federico Retyk and
                  Rus Poves and
                  Juan Aizpuru and
                  Hermenegildo Fabregat and
                  Vaidotas Simkus and
                  Em{\'{\i}}lia Garcia Casademont},
  title        = {Learning Job Titles Similarity from Noisy Skill Labels},
  journal      = {CoRR},
  volume       = {abs/2207.00494},
  year         = {2022},
  url          = {https://doi.org/10.48550/arXiv.2207.00494},
  doi          = {10.48550/ARXIV.2207.00494},
  eprinttype    = {arXiv},
  eprint       = {2207.00494},
  bibsource    = {dblp computer science bibliography, https://dblp.org}
}

@inproceedings{retyk2023r,
  author       = {Federico Retyk and
                  Hermenegildo Fabregat and
                  Juan Aizpuru and
                  Mariana Taglio and
                  Rabih Zbib},
  title        = {R{\'{e}}sum{\'{e}} Parsing as Hierarchical Sequence Labeling:
                  An Empirical Study},
  booktitle    = {Proceedings of the 3rd Workshop on Recommender Systems for Human Resources
                  (RecSys in {HR} 2023) co-located with the 17th {ACM} Conference on
                  Recommender Systems (RecSys 2023), Singapore, Singapore, 18th-22nd
                  September 2023},
  series       = {{CEUR} Workshop Proceedings},
  volume       = {3490},
  publisher    = {CEUR-WS.org},
  year         = {2023},
  url          = {https://ceur-ws.org/Vol-3490/RecSysHR2023-paper\_10.pdf},
  bibsource    = {dblp computer science bibliography, https://dblp.org}
}

@article{garcia2025measuring,
  title={Measuring Gender Bias in Job Title Matching for Grammatical Gender Languages},
  author={Garc{\'\i}a-Sardi{\~n}a, Laura and Fabregat, Hermenegildo and Deniz, Daniel and Zbib, Rabih},
  journal={arXiv preprint arXiv:2509.13803},
  year={2025}
}

@article{garcia2023normalisation,
  title={Normalisation of Education Information in Digitalised Recruitment Processes},
  author={Garc{\'\i}a-Sardi{\~n}a, Laura and Retyk, Federico and Fabregat, Hermenegildo and Alvarez Lacasa, Lucas and Poves, Rus and Zbib, Rabih},
  journal={Procesamiento del Lenguaje Natural},
  year={2023},
  publisher={Sociedad Espa{\~n}ola para el Procesamiento del Lenguaje Natural},
  volume={71},
  pages={63--73},
	issn = {1989-7553}
}

@inproceedings{deniz2024combined,
  author       = {Daniel Deniz and
                  Federico Retyk and
                  Laura Garc{\'{\i}}a{-}Sardi{\~{n}}a and
                  Hermenegildo Fabregat and
                  Luis Gasc{\'{o}} and
                  Rabih Zbib},
  editor       = {Mesut Kaya and
                  Toine Bogers and
                  David Graus and
                  Chris Johnson and
                  Jens{-}Joris Decorte and
                  Tijl De Bie},
  title        = {Combined Unsupervised and Contrastive Learning for Multilingual Job
                  Recommendation},
  booktitle    = {Proceedings of the 4th Workshop on Recommender Systems for Human Resources
                  (RecSys-in-HR 2024) co-located with the 18th {ACM} Conference on Recommender
                  Systems (RecSys 2024), Bari, Italy, 14th-18th October 2024},
  series       = {{CEUR} Workshop Proceedings},
  volume       = {3788},
  publisher    = {CEUR-WS.org},
  year         = {2024},
  url          = {https://ceur-ws.org/Vol-3788/RecSysHR2024-paper\_3.pdf},
  bibsource    = {dblp computer science bibliography, https://dblp.org}
}

@inproceedings{retyk2024melo,
  author       = {Federico Retyk and
                  Luis Gasc{\'{o}} and
                  Casimiro Pio Carrino and
                  Daniel Deniz and
                  Rabih Zbib},
  editor       = {Mesut Kaya and
                  Toine Bogers and
                  David Graus and
                  Chris Johnson and
                  Jens{-}Joris Decorte and
                  Tijl De Bie},
  title        = {{MELO:} An Evaluation Benchmark for Multilingual Entity Linking of
                  Occupations},
  booktitle    = {Proceedings of the 4th Workshop on Recommender Systems for Human Resources
                  (RecSys-in-HR 2024) co-located with the 18th {ACM} Conference on Recommender
                  Systems (RecSys 2024), Bari, Italy, 14th-18th October 2024},
  series       = {{CEUR} Workshop Proceedings},
  volume       = {3788},
  publisher    = {CEUR-WS.org},
  year         = {2024},
  url          = {https://ceur-ws.org/Vol-3788/RecSysHR2024-paper\_2.pdf},
  bibsource    = {dblp computer science bibliography, https://dblp.org}
}

@inproceedings{bogers2025fifth,
  title={Fifth Workshop on Recommender Systems for Human Resources (RecSys in HR 2025)},
  author={Bogers, Toine and Kaya, Mesut and Decorte, Jens-Joris and Johnson, Chris and Bied, Guillaume},
  booktitle={Proceedings of the Nineteenth ACM Conference on Recommender Systems},
  pages={1373--1377},
  year={2025}
}

@article{otani2024natural,
  title={Natural Language Processing for Human Resources: A Survey},
  author={Otani, Naoki and Bhutani, Nikita and Hruschka, Estevam},
  journal={arXiv preprint arXiv:2410.16498},
  year={2024}
}

@inproceedings{lo2025ai,
  title={AI hiring with llms: A context-aware and explainable multi-agent framework for resume screening},
  author={Lo, Frank P-W and Qiu, Jianing and Wang, Zeyu and Yu, Haibao and Chen, Yeming and Zhang, Gao and Lo, Benny},
  booktitle={Proceedings of the Computer Vision and Pattern Recognition Conference},
  pages={4184--4193},
  year={2025}
}

@article{gan2024application,
  title={Application of LLM agents in recruitment: A novel framework for resume screening. arXiv},
  author={Gan, C and Zhang, Q and Mori, T},
  journal={arXiv preprint arXiv:2401.08315},
  year={2024}
}

@article{zhao2021embedding,
  title={Embedding-based recommender system for job to candidate matching on scale},
  author={Zhao, Jing and Wang, Jingya and Sigdel, Madhav and Zhang, Bopeng and Hoang, Phuong and Liu, Mengshu and Korayem, Mohammed},
  journal={arXiv preprint arXiv:2107.00221},
  year={2021}
}

@article{decorte2021jobbert,
  author       = {Jens{-}Joris Decorte and
                  Jeroen Van Hautte and
                  Thomas Demeester and
                  Chris Develder},
  title        = {JobBERT: Understanding Job Titles through Skills},
  journal      = {CoRR},
  volume       = {abs/2109.09605},
  year         = {2021},
  url          = {https://arxiv.org/abs/2109.09605},
  eprinttype    = {arXiv},
  eprint       = {2109.09605},
  bibsource    = {dblp computer science bibliography, https://dblp.org}
}

@inproceedings{senger2024deep,
  title={Deep Learning-based Computational Job Market Analysis: A Survey on Skill Extraction and Classification from Job Postings},
  author={Senger, Elena and Zhang, Mike and van der Goot, Rob and Plank, Barbara},
  year={2024},
booktitle = "Proceedings of the First Workshop on Natural Language Processing for Human Resources (NLP4HR 2024)",
    address = {"St. Julian{'}s, Malta"},
    publisher = {"Association for Computational Linguistics"},
    url = "https://aclanthology.org/2024.nlp4hr-1.1/",
    pages = "1--15"
}

@inproceedings{zhang2022skill,
  author       = {Mike Zhang and
                  Kristian N{\o}rgaard Jensen and
                  Rob van der Goot and
                  Barbara Plank},
  editor       = {Mesut Kaya and
                  Toine Bogers and
                  David Graus and
                  Sepideh Mesbah and
                  Chris Johnson and
                  Francisco Guti{\'{e}}rrez},
  title        = {Skill Extraction from Job Postings using Weak Supervision},
  booktitle    = {Proceedings of the 2nd Workshop on Recommender Systems for Human Resources
                  (RecSys-in-HR 2022) co-located with the 16th {ACM} Conference on Recommender
                  Systems (RecSys 2022), Seattle, USA, 18th-23rd September 2022},
  series       = {{CEUR} Workshop Proceedings},
  volume       = {3218},
  publisher    = {CEUR-WS.org},
  year         = {2022},
  url          = {https://ceur-ws.org/Vol-3218/RecSysHR2022-paper\_10.pdf},
  bibsource    = {dblp computer science bibliography, https://dblp.org}
}

@inproceedings{zhang2022skillspan,
  author       = {Mike Zhang and
                  Kristian N{\o}rgaard Jensen and
                  Sif Dam Sonniks and
                  Barbara Plank},
  editor       = {Marine Carpuat and
                  Marie{-}Catherine de Marneffe and
                  Iv{\'{a}}n Vladimir Meza Ru{\'{\i}}z},
  title        = {SkillSpan: Hard and Soft Skill Extraction from English Job Postings},
  booktitle    = {Proceedings of the 2022 Conference of the North American Chapter of
                  the Association for Computational Linguistics: Human Language Technologies,
                  {NAACL} 2022, Seattle, WA, United States, July 10-15, 2022},
  pages        = {4962--4984},
  publisher    = {Association for Computational Linguistics},
  year         = {2022},
  url          = {https://doi.org/10.18653/v1/2022.naacl-main.366},
  doi          = {10.18653/V1/2022.NAACL-MAIN.366},
  bibsource    = {dblp computer science bibliography, https://dblp.org}
}

@article{10.1145/1852102.1852106,
author = {Webber, William and Moffat, Alistair and Zobel, Justin},
title = {A similarity measure for indefinite rankings},
year = {2010},
issue_date = {November 2010},
publisher = {Association for Computing Machinery},
address = {New York, NY, USA},
volume = {28},
number = {4},
issn = {1046-8188},
url = {https://doi.org/10.1145/1852102.1852106},
doi = {10.1145/1852102.1852106},
journal = {ACM Trans. Inf. Syst.},
month = nov,
articleno = {20},
numpages = {38}
}

@inproceedings{hruschka2024proceedings,
  title={Proceedings of the First Workshop on Natural Language Processing for Human Resources (NLP4HR 2024)},
  author={Hruschka, Estevam and Lake, Thom and Otani, Naoki and Mitchell, Tom},
  booktitle={Proceedings of the First Workshop on Natural Language Processing for Human Resources (NLP4HR 2024)},
  year={2024},
  publisher = {"Association for Computational Linguistics"},
  url = "https://aclanthology.org/2024.nlp4hr-1.0/"
}

\end{document}